\documentclass[format=sigconf,anonymous=false,review=false]{acmart}
\usepackage[dvipsnames]{xcolor}
\AtBeginDocument{%
  }
\usepackage{caption}
\usepackage{float}
\usepackage{algorithm}
\usepackage{algpseudocode}
\setcopyright{acmlicensed}
\usepackage{amsmath}
\usepackage{array}

\setcopyright{none}
\renewcommand\footnotetextcopyrightpermission[1]{}
\acmConference{}{}{}
\acmBooktitle{}
\acmISBN{}
\acmDOI{}
\acmYear{}
\acmPrice{}
\begin{document}

\title{Evolving Multi-Channel Confidence-Aware Activation Functions for Missing Data with Channel Propagation}

% 1st author (First author)
\author{Naeem Shahabi Sani}
\email{shahabi@ou.edu}
\affiliation{%
  \institution{University of Oklahoma}
  \city{Norman}
  \state{Oklahoma}
  \country{USA}
}

% 2nd author
\author{Ferial Najiantabriz}
\email{ferial@ou.edu}
\affiliation{%
  \institution{University of Oklahoma}
  \city{Norman}
  \state{Oklahoma}
  \country{USA}
}

% 3rd author
\author{Shayan Shafaei}
\email{shayan.shafaei@ou.edu}
\affiliation{%
  \institution{University of Oklahoma}
  \city{Norman}
  \state{Oklahoma}
  \country{USA}
}

% 4th author (Senior/last author)
\author{Dean F. Hougen}
\email{hougen@ou.edu}
\affiliation{%
  \institution{University of Oklahoma}
  \city{Norman}
  \state{Oklahoma}
  \country{USA}
}

\begin{abstract}
Learning in the presence of missing data can result in biased predictions and poor generalizability, among other difficulties, which data imputation methods only partially address. In neural networks, activation functions significantly affect performance, yet typical options (e.g., ReLU, Swish) operate only on feature values and do not account for missingness indicators or confidence scores. We propose \emph{Three-Channel Evolved Activations} (3C-EA), which we evolve using Genetic Programming to produce multivariate activation functions $f(x,m,c)$ in the form of trees that take (i) the feature value $x$, (ii) a missingness indicator $m$, and (iii) an imputation confidence score $c$. To make these activations useful beyond the input layer, we introduce \emph{ChannelProp}, an algorithm that deterministically propagates missingness and confidence values via linear layers based on weight magnitudes, retaining reliability signals throughout the network. We evaluate 3C-EA and ChannelProp on datasets with natural and injected (Missing Completely at Random, Missing at Random, and Missing Not at Random) missingness at multiple rates under identical preprocessing and splits. Results indicate that integrating missingness and confidence inputs into the activation search improves classification performance under missingness.
\end{abstract}

\begin{comment}
Activation functions significantly affect neural network performance; yet, typical options (e.g., ReLU, Swish) utilize fixed scalar nonlinearities that operate only on the feature value $x$ and do not condition on missingness indicators or confidence scores. This constraints is crucial in the context of missing data, as networks frequently receive imputed values without an obvious way for identifying suspicious inputs from observed ones. We propose \emph{Three-Channel Evolutionary Activations}, which use genetic programming to evolve multivariate activation trees $f(x,m,c)$ that take (i) the feature value $x$, (ii) a missingness indicator $m$, and (iii) an imputation-confidence score $c$. To make these activations useful beyond the input layer, we introduce ChannelProp, an algorithm with determinism that distributes missingness and confidence via linear layers based on weight magnitudes, keeping reliability signals throughout the network. We evaluate on datasets with natural and injected (MCAR/MAR/MNAR) missingness at multiple rates under identical preprocessing and splits; results indicate that integrating missingness and confidence inputs into the activation search improves classification performance under missingness.
\end{comment}


\maketitle

\section{Introduction}
In real-world machine learning applications, data is rarely complete~\cite{little2019statistical}. Missing values are common in fields such as medical informatics, finance, and sensor networks, requiring preprocessing techniques such as imputation or masking~\cite{little2019statistical,schafer2002missing,batista2003analysis}. Although these strategies allow models to work with incomplete information, they frequently obscure an essential contrast between observed data and imputed projections~\cite{van2012flexible}. Traditional neural networks intensify this problem by depending on fixed scalar activation functions (e.g., ReLU, Swish, ELU) that function only on the feature value $x$ and fail to clearly distinguish between high-confidence observations and lower confidence imputations.

A common strategy for addressing missing data is to use missingness masks or indicators as secondary input features. This enables the network to use missingness information; nevertheless, the nonlinear transformations remain unaltered, and any reliability information must be inferred implicitly via linear weights~\cite{bingham2020evolutionary}. Consequently, designing nonlinear behaviors that explicitly respond to data quality remains a significant challenge.

Neuroevolution provides a structured solution by allowing the automatic identification of neural components that are challenging to design by hand~\cite{risi-et-al:book2025}. Previous research has demonstrated that evolutionary search may identify enhanced scalar activation functions $f(x)$ and effectively navigate complex function spaces via genetic programming (GP)~\cite{koza1994genetic,bingham2020evolutionary}. Nonetheless, current activation function searches are primarily limited to scalar inputs and are unable to directly utilize data-quality information. This work presents \emph{Three-Channel Evolved Activations} (3C-EA), which expands activation evolution to multivariate computation trees represented as $f(x,m,c)$, where $x$ is the feature value, $m$ acts as a binary missingness indicator, and $c$ shows an imputation confidence score. In our approach, confidence is set to $1.0$ for observed values and decreased for imputed values according to feature-specific missingness rates derived from the training data.

To improve multi-channel activations in deep networks, we present \emph{ChannelProp}, a deterministic propagation rule that communicates missingness and confidence via linear layers by employing normalized absolute weight values. This guarantees that reliability information gets transformed together with feature values instead of being eliminated post-input layer. Accordingly, evolved activations can locally modify their responses, such as by diminishing activations in low-confidence scenarios or altering regimes in the absence of inputs. We evaluate the proposed method on datasets with natural missing values and on initially full datasets with intentionally created missingness under missing completely at random, missing at random, and missing not at random (MCAR, MAR, and MNAR, respectively) methods~\cite{little2019statistical}. Performance is evaluated against standard fixed activation baselines (ReLU, Swish, LeakyReLU, ELU) with identical preprocessing procedures and identical train, validation, and test splits.

\noindent\textbf{Contributions.}
(i) We present a GP-based system for evolving multivariate activation functions that clearly depend on feature values, missingness indicators, and confidence scores;
(ii) We introduce ChannelProp, a deterministic approach to propagating reliability information across linear layers;
(iii) We illustrate via extensive experiments that integrating data-quality information into the activation search space results in measurable performance enhancements, especially in scenarios of structured and value-dependent missingness.

\section{Related Work}
Our work bridges the gap between work in handling missing data (\S\ref{sec:missing}) and the evolution of neural activation functions (\S\ref{sec:neuroevo}).

\subsection{Missing Data Handling} 
\label{sec:missing}
Missing data has been widely studied in machine learning~\cite{jerez2010missing,van2011mice,yoon2018gain,nazabal2020handling,sun2023deep,emmanuel2021survey}. Traditional methods use single or multiple imputation based on assumptions such as MCAR, MAR, and MNAR~\cite{rubin1976inference,little2019statistical}. Mean imputation and model-based methods like expectation-maximization and k-nearest neighbors are common strategies. They are fast to run but may hide uncertainty~\cite{schafer2002missing,batista2003analysis}. Recent research integrates missingness markers as supplementary inputs, enabling models to condition on observedness data~\cite{lipton2016directly,che2018recurrent}. Evolutionary methods have been investigated for missing-data issues, primarily via imputation-based optimization~\cite{aydilek2013hybrid,lobato2015evolutionary}. Nevertheless, most current methods consider activation functions as fixed factors, constraining the capacity of nonlinear transformations to directly adjust to data dependability~\cite{bingham2020evolutionary,apicella2021survey,dubey2022activation}.

\subsection{Evolved Neural Activation Functions}
\label{sec:neuroevo}
Activation functions are a crucial part of neural network performance~\cite{dubey2022activation, ramachandran2017searching, glorot2010understanding}. Although researchers have long depended on fixed nonlinearities, the trend has been towards automated discovery to move past the shortcomings of manually designed functions~\cite{kunc2024three}.

Ramachandran et al.~\cite{ramachandran2017searching} applied reinforcement learning to automatically identify enhanced scalar activation functions, whereas later researchers have employed evolutionary computation to expand the search space. Bingham et al.~\cite{bingham2020evolutionary} used GP to evolve scalar activation functions in a tree-based search space, showing improvements over ReLU, while Parisi et al.~\cite{parisi2024optimal} used genetic algorithms to optimize nonlinear operators in a predefined activation search space.

%Nader and Azar~\cite{nader2020searching} proposed a self-adaptive evolutionary algorithm. Unlike traditional methods that use fixed mutation and crossover probabilities, this algorithm has the ability to change its hyperparameters during the search process, thus optimizing the discovery process. 

Our method expands on the evolutionary line of research by including the discovery of evolving activation functions from scalar nonlinearities to multivariate, reliability-aware activations that work together on feature values, missingness indicators, and confidence signals.

\section{Method}
\label{sec:method}
This section explains the three-channel input representation 
(\S\ref{sec:method_setup}), describes how the imputation, missingness, and confidence channels are constructed 
(\S\ref{sec:channels}), introduces ChannelProp for propagating reliability metadata (\S\ref{sec:channelprop}), 
and details the fitness evaluation (\S\ref{sec:fitness}) and evolutionary search process (\S\ref{sec:gp}).
\subsection{Problem Setup and Three-Channel Inputs}
\label{sec:method_setup}
Let $\mathcal{D}=\{(\mathbf{x}^{(i)}, y^{(i)})\}_{i=1}^{n}$ represent a tabular classification dataset comprised of $d$ features, where $\mathbf{x}^{(i)}\in\mathbb{R}^{d}$ may include missing values and $y^{(i)}$ represents the class label. Standard multilayer perceptrons (MLPs) often employ a fixed scalar nonlinearity $g(z)$ (e.g., ReLU, Swish, ELU) for each pre-activation $z$, treating both observed values and imputed estimations similarly.

We present explicit reliability information to the nonlinearity by creating three channels for each feature: (i) an imputed value channel $\tilde{x}$, (ii) a missingness indicator $m\in[0,1]$, and (iii) a confidence score $c\in[0,1]$. In the input layer, $m\in\{0,1\}$ indicates the absence of a feature; in subsequent layers, $m$ transforms into a soft missingness probability through deterministic propagation (Section~\ref{sec:channelprop}). Our purpose is to create a unique multivariate activation function 
\begin{equation}
f(\tilde{x}, m, c),
\end{equation}
through GP, and to apply this activation in each hidden layer of a three-channel MLP (3C-MLP). This enables the nonlinearity to adjust responses according to the degree of missing data and reliability. Standard numerical safeguards are used throughout this section as common implementation details.

\subsection{Channel Construction: Imputation, Missingness, and Confidence}
\label{sec:channels}

Each dataset handles missing values via an intentional three-channel representation constructed before network training. Let $\mathbf{X}\in\mathbb{R}^{n\times d}$ represent the feature matrix after missingness injection (when applicable). We derive the following channels.

\begin{description}
\item[Imputed Value Channel.]
Missing entries are imputed using feature-specific means calculated only from the training set. Let $\mu_j$ represent the mean of feature $j$ calculated from the observed training data. All missing values in the training, validation, and test subsets are substituted with $\mu_j$. This approach prevents data leakage while ensuring a uniform numeric input $\tilde{x}$ for the next processing step.

\item[Missingness Channel.]
A binary missingness mask $m$ is generated concurrently with imputation, where $m_{ij}=1$ indicates that feature $j$ of sample $i$ is missing, while $m_{ij}=0$ indicates its presence. Although $m$ is binary at the input layer, it turns into a soft missingness probability in subsequent layers via deterministic propagation (Section~\ref{sec:channelprop}).

\item[Confidence Channel.]
To quantify the reliability of imputed values, we provide a confidence score $c\in[0,1]$ to each feature. For observed entries, confidence is determined as $c_{ij}=1.0$. For missing entries, confidence decreases according to the feature-specific missing rate derived from the training data. Let \[ r_j = \frac{1}{n_{\text{train}}}\sum_{i=1}^{n_{\text{train}}} m_{ij} \] represent the proportion of missing values in feature $j$ inside the training dataset. 
Confidence is defined as
\[
c_{ij}=
\begin{cases}
1.0, & \text{if } m_{ij}=0,\\
\max(\tau,\,1-r_j), & \text{if } m_{ij}=1,
\end{cases}
\]
so imputed values from rarely missing features receive higher confidence than those from frequently missing ones. Here, $\tau=0.1$ is a lower bound that avoids degenerate near-zero confidence values.
This creates a dataset-driven, feature-specific confidence heuristic that indicates the relative reliability of imputed values without adding extra learned factors.

\end{description}

\subsection{ChannelProp: Deterministic Propagation of Reliability Metadata}
\label{sec:channelprop}

To enable reliability-aware activations beyond the input layer, missingness and confidence information must be propagated through linear transformations. Standard mask-as-feature approaches provide no such mechanism, causing reliability metadata to vanish after the first layer. We address this limitation with \emph{ChannelProp}, a deterministic propagation rule that transports missingness and confidence alongside feature values across linear layers, using the linear layer's weights to determine how strongly each output depends on inputs of differing reliability.

Let $\mathbf{c}_{\text{in}} \in [0,1]^{d_{\text{in}}}$ and
$\mathbf{m}_{\text{in}} \in [0,1]^{d_{\text{in}}}$ denote the confidence and
missingness vectors associated with the input features of a linear layer,
and let $\mathbf{c}_{\text{out}} \in [0,1]^{d_{\text{out}}}$ and
$\mathbf{m}_{\text{out}} \in [0,1]^{d_{\text{out}}}$ denote the corresponding
quantities for the output neurons. We define the observedness vector as
$\mathbf{o}_{\text{in}} = 1 - \mathbf{m}_{\text{in}}$.
All channel vectors are treated as row vectors.

Consider a linear transformation with weight matrix
$W \in \mathbb{R}^{d_{\text{out}} \times d_{\text{in}}}$.
ChannelProp constructs a non-negative routing matrix
\[
A = |W| + \varepsilon,
\]
where $|\cdot|$ denotes elementwise absolute value and $\varepsilon$ is a small constant used only to avoid division by zero during normalization. Row-normalization yields
\[
\tilde{A}_{ij} = \frac{A_{ij}}{\sum_k A_{ik}}.
\]

Confidence and observedness are propagated as weighted mixtures
\[
\mathbf{c}_{\text{out}} = \mathbf{c}_{\text{in}} \tilde{A}^{\top},
\qquad
\mathbf{o}_{\text{out}} = \mathbf{o}_{\text{in}} \tilde{A}^{\top},
\]
and the output missingness channel is defined as
\[
\mathbf{m}_{\text{out}} = 1 - \mathbf{o}_{\text{out}}.
\]
All propagated signals are clipped to $[0,1]$.

ChannelProp introduces deterministic parameters and depends solely on the linear
weights. While missingness is binary at the input layer, the propagated channel
becomes a soft reliability signal reflecting partial dependence on missing
inputs. This allows evolved activations to modulate their behavior continuously
based on propagated reliability information.

\subsection{Fitness Evaluation}
\label{sec:fitness}

The fitness of a candidate activation tree $T$ is assessed by embedding it into a 3C-MLP and evaluating its predictive performance for a constrained training period. This short horizon evaluation enables efficient exploration of the activation search space while preserving a meaningful correlation between fitness and generalization performance.

\subsubsection{Evaluation protocol.}
For each candidate activation, we construct a 3C-MLP in which all hidden layers use the same evolved activation $T$. The network is trained on the training split using the Adam optimizer \cite{KingmaB14} with a fixed  learning rate of $0.001$ and weight decay of $1 \times 10^{-4}$, and early stopping is applied with a small patience to limit overfitting and computational cost. Each GP run evaluates 3{,}000 candidate activations, so we use a reduced training schedule to keep the search feasible. Validation accuracy has been tracked during training, and the best validation accuracy $A_{\mathrm{val}}(T)$ detected during this process is stored.

\subsubsection{Fitness function.}
The overall fitness of a tree $T$ is defined as
\begin{equation}
\label{eq:fitness}
F(T) = A_{\mathrm{val}}(T)
+ \lambda_d D(T)
- \lambda_s N(T)
- \lambda_h \big(H(T) - 1\big),
\end{equation}
where $D(T)\in\{0,1,2,3\}$ counts the number of distinct input channels ($\tilde{x}$, $m$, $c$) referenced by the tree, $N(T)$ is the number of nodes, and $H(T)$ is the tree depth. Trees of depth $H=1$ are assigned zero fitness because they correspond only to terminal nodes, and excluding them avoids spending expensive fitness evaluations on trivial candidates that fall outside the intended search space. The coefficient $\lambda_d$ mildly encourages use of the reliability channels, while $\lambda_s$ and $\lambda_h$ penalize size and depth respectively to control code bloat, since trees with equal node counts can differ significantly in nesting depth. In all experiments, these coefficients are fixed to $(\lambda_d, \lambda_s, \lambda_h) = (0.01, 0.0001, 0.0002)$.

\subsection{GP Representation and Search}
\label{sec:gp}

\begin{figure}[htbp]
    \centering
        \vspace{-4ex}
        \includegraphics[width=0.5\columnwidth]{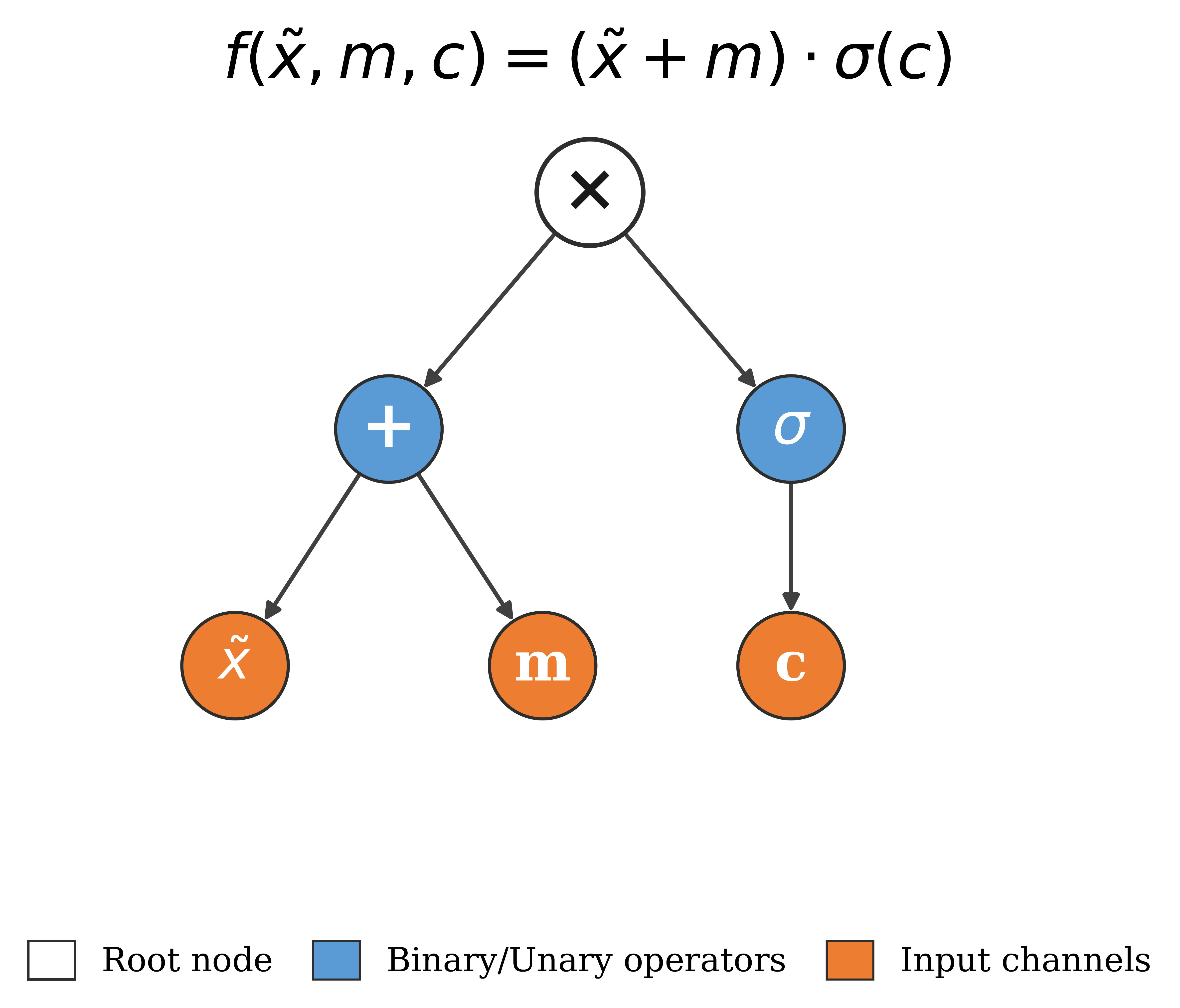}
    \vspace{-2ex}
    \caption{\small Tree representation of an evolved three-channel activation function.}
    \Description{    }
    \label{fig:evolved}
\end{figure}

Each candidate activation function is represented as a symbolic expression tree defining a multivariate nonlinearity
$f(\tilde{x}, m, c)$, where $\tilde{x}$ denotes the imputed feature value, $m$ the missingness indicator,
and $c$ the propagated confidence score. An example tree is shown in Figure \ref{fig:evolved}. This representation allows evolved activations to condition explicitly on
reliability metadata.

\subsubsection{Terminal and function sets.}
The terminal set consists of the three input channels $\{\tilde{x}, m, c\}$ and a fixed set of real-valued constants
$\mathcal{C}=\{0,\pm0.1,\pm0.5,\pm1,\pm2\}$. The function set includes unary operators (e.g., $\tanh$, logistic sigmoid aka $\sigma$, ReLU,
softplus, $\exp$, $\log|\cdot|$, square, absolute value) and binary operators (addition, subtraction, multiplication,
protected division with a small $\varepsilon$ added when $|x_{2}|$ is close to zero \cite{koza1994genetic}, $\min$, and $\max$). All operators are applied elementwise with numerical safeguards to ensure
stability. Complete operator lists are provided in Tables~\ref{tab:unary-ops} and~\ref{tab:binary-ops}.

\begin{table}[thbp]
\centering
\renewcommand{\arraystretch}{0.78}
\setlength{\abovecaptionskip}{2pt}
\setlength{\belowcaptionskip}{0pt}
\caption{\small Unary operators available to the GP search.}
\label{tab:unary-ops}
\footnotesize
\begin{tabular}{llll}
\toprule
\multicolumn{4}{c}{\textsc{Unary Operations}} \\
\midrule
$x$                        & $-x$                         & $|x|$                        & $x^{2}$ \\
$x^{3}$                    & $\sqrt{x}$   & $\exp(x)$                   & $\log(|x|)$ \\
$\sin(x)$                  & $\cos(x)$                    & $\tanh(x)$                  & $\sigma(x)$\\
$\operatorname{ReLU}(x)$   & $\operatorname{LeakyReLU}(x)$ & $\operatorname{ELU}(x)$     & $\operatorname{softplus}(x)$ \\
\bottomrule
\end{tabular}
\end{table}

\begin{table}[thbp]
\centering
\caption{\small Binary operators available to the GP search.}
\label{tab:binary-ops}
\begin{small}
\begin{tabular}{lll}
\toprule
\multicolumn{3}{c}{\textsc{Binary Operations}} \\
\midrule
$x_{1} + x_{2}$ & $x_{1} - x_{2}$ & $x_{1} \cdot x_{2}$ \\
$x_{1} / x_{2}$ & $\max(x_{1}, x_{2})$ & $\min(x_{1}, x_{2})$ \\
\bottomrule
\end{tabular}
\end{small}
\end{table}

\subsubsection{Evolutionary search.}
Activation trees are evolved using a generational GP algorithm with fitness-proportional selection,
subtree crossover, and mutation. To control excessive growth and maintain interpretability, trees are constrained to a
maximum depth throughout initialization and variation, and elitism preserves the top-performing individuals across
generations. The algorithm is found in Algorithm~\ref{alg:3cea}. Evolutionary hyperparameters are fixed across all experiments and summarized in
Table~\ref{tab:hyperparams}.

\begin{algorithm}[thbp]
\caption{\small Evolution of three-channel activation functions (3C-EA)}
\label{alg:3cea}
\scriptsize
\begin{algorithmic}[1]
\Require Dataset $\mathcal{D}$ with missing values, GP parameters
\Ensure Best evolved activation $T^*$

\State Construct three-channel inputs $(\tilde{x}, m, c)$ using mean imputation, missingness masks, and feature-wise confidence
\State Initialize population $\mathcal{P}_0$ of activation trees with bounded depth

\For{$g = 1$ to $G$}
    \ForAll{$T \in \mathcal{P}_{g-1}$}
        \State Embed $T$ into a three-channel MLP
        \State Train network under short horizon with early stopping
        \State Evaluate fitness $F(T)$ using validation accuracy and complexity penalties
    \EndFor
    \State Select parents via softmax fitness-proportional sampling
    \State Generate offspring via subtree crossover and mutation
    \State Preserve top $k$ individuals by elitism
    \State Form next population $\mathcal{P}_g$
\EndFor

\State \Return $T^* = \arg\max_{T \in \bigcup_{g=0}^{G} \mathcal{P}_g} F(T)$
\end{algorithmic}
\end{algorithm}

\begin{table}[thbp]
\centering
\renewcommand{\arraystretch}{0.78}
\setlength{\abovecaptionskip}{2pt}
\setlength{\belowcaptionskip}{0pt}
\caption{\small GP hyperparameters.}
\label{tab:hyperparams}
\footnotesize
\begin{tabular}{lc}
\toprule
\textbf{Parameter} & \textbf{Value} \\
\midrule
Population size & 100 \\
Generations & 30 \\
Maximum tree depth & 3 \\
Crossover probability & 0.7 \\
Mutation probability & 0.15 \\
Elite size & 2 \\
\bottomrule
\end{tabular}
\end{table}

\section{Experiments, Results, and Discussion}

We present our experimental setup, results on datasets with naturally missing data, results on datasets that are originally complete but from which we synthetically remove data to carefully study various kinds of missing data, results of an ablation study to demonstrate the contribution of each component of our proposed method, and an analysis of the evolved activation functions.

\subsection{Experimental Setup}
We describe the datasets used (\S\ref{sec:datasets}), preprocessing  steps (\S\ref{sec:preprocessing}), evaluation metrics and protocol  (\S\ref{sec:metrics}), and baselines and implementation details  (\S\ref{sec:baselines}).
\subsubsection{Datasets}
\label{sec:datasets}
We evaluate our method using two different types of datasets from the UCI Machine Learning Repository~\cite{uci}. Dataset details and information about missing data are presented in Table~\ref{tab:datasets}. This experimental design provides the evaluation of model performance under both naturally existing and intentionally generated missing data scenarios. The datasets include Hepatitis, HouseVotes84, Credit Approval, Mammographic, Cylinder Bands, Heart Disease, and Adult. The datasets reflect significant variation in size, feature dimensionality, feature type, and the level of missing data, with the proportion of incomplete instances varying from moderate to severe. This diversity offers a practical framework to evaluate robustness in the presence of naturally partial data.

The second group includes datasets that are naturally complete and free of missing values, including Mushroom, WDBC, PIMA, Sonar, and Glass. In these datasets, missingness is intentionally generated in a controlled manner to help with systematic study under standardized conditions. The following section provides details on missingness mechanisms and injection methods.
\subsubsection{Preprocessing}
\label{sec:preprocessing}

For all datasets, missing values are imputed using mean imputation generated entirely from the training data. A binary signal indicating missingness is generated to show whether each feature was initially observed or missing together with the values of the imputed feature. An imputation-confidence score is computed for each feature entry, starting at 1 for observed features and decreasing for imputed entries according to the missingness frequency derived from the training data. This enables the proposed technique to differentiate between the observed and imputed features.
\subsubsection{Evaluation Metrics and Experimental Protocol}
\label{sec:metrics}
We use Accuracy, Precision, Recall (Sensitivity), Specificity, F1-score, and Area Under the ROC Curve (AUC) as the metrics to evaluate the performance of the classification task~\cite{powers2011evaluation,hand2001simple}.
In order to make our results robust and reproducible, all experiments are performed with 30 different runs and random seeds. For each metric, we calculate the mean and standard deviation. Statistical significance is assessed using a one-sided Wilcoxon signed-rank test~\cite{derrac2011practical} on paired per-run results, with alternative hypothesis $H_1$: 3C-EA $>$ baseline. Symbols indicate significance against ReLU ($^{*}$/$^{**}$), Swish ($^{\dagger}$/$^{\dagger\dagger}$), LeakyReLU ($^{\ddagger}$/$^{\ddagger\ddagger}$), and ELU ($^{\S}$/$^{\S\S}$), for $p<0.05$ and $p<0.01$, respectively.

\subsubsection{Baselines and Implementation}
\label{sec:baselines}
We evaluate 3C-EA against standard fixed activation functions including ReLU, Swish, Leaky ReLU, and ELU. Baseline models use neural network architectures, optimization strategy, preprocessing pipeline, and data partitions identical to the proposed method, differing only in the selection of the activation function. This guarantees that performance divergence is entirely related to the activation design, without influence from other factors. The proposed 3C-EA approach has been evaluated using a 3C-MLP architecture, in which each nonlinear layer employs the same evolved activation function  $f(\tilde{x}, m, c)$. Activation functions are specially constructed for each dataset by GP, as detailed in Table \ref{tab:hyperparams}. After evolution, the selected activation is retrained for final evaluation for up to 100 epochs with patience 15. All experiments were implemented in Python using PyTorch and scikit-learn.

\begin{table}[t]
\centering
\caption{Datasets used in the experiments.}
\label{tab:datasets}
\footnotesize
\renewcommand{\arraystretch}{0.80}
\setlength{\tabcolsep}{3pt}
\setlength{\abovecaptionskip}{2pt}
\setlength{\belowcaptionskip}{0pt}
\begin{tabular}{@{}lccccc@{}}
\toprule
Name & Inst. & Features & Feature Type & Classes & Incomplete (\%) \\
\midrule
Hepatitis        & 155  & 19 & Mixed & 2 & 48.39 \\
HouseVotes84     & 435  & 16 & Cat.   & 2 & 46.67 \\
Credit Approval  & 690  & 14 & Mixed & 2 & 4.20  \\
Mammographic & 961 & 5 & Mixed & 2 & 13.53\\
Cylinder Bands   & 512  & 39 & Mixed & 2 & 38.0 \\
Heart Disease    & 303  & 13 & Mixed & 2 & 6.6  \\
Adult             & 48842 & 14 & Mixed & 2 & 7.4 \\
Mushroom         & 8124 & 22 & Cat.   & 2 & 0     \\
WDBC             & 569  & 30 & Num.   & 2 & 0     \\
PIMA             & 768  & 8  & Num.   & 2 & 0     \\
Sonar            & 208  & 60 & Num.   & 2 & 0     \\
Glass            & 214  & 9  & Num.   & 6 & 0     \\
\bottomrule
\end{tabular}
\end{table}

\subsection{Results on Real World Incomplete Datasets}

\begin{table*}[htbp]
\caption{\small Performance Comparison on Real World Incomplete datasets}
\renewcommand{\arraystretch}{0.72}
\label{tab:Incomplete}
\centering
\scriptsize
\setlength{\tabcolsep}{3pt}
\setlength{\abovecaptionskip}{2pt}
\setlength{\belowcaptionskip}{0pt}
\begin{tabular}{lccccccc}
\toprule
Dataset & Method & TestAcc & Prec & Rec & Spec & F1 & AUC \\
\midrule

Hepatitis 
  &3C-EA   & \textbf{0.8039$\pm$0.0407}
 & $0.8478\pm0.0748^{**\ddagger\ddagger}$ & $0.9283\pm0.0846^{\dagger\S}$ & $0.3771\pm0.3449^{**\ddagger\ddagger}$ & \textbf{0.8796$\pm$0.0263} & $0.7195\pm0.2005^{*\ddagger\ddagger}$ \\
&ReLU      & 0.7703$\pm$0.0550 & 0.8048$\pm$0.0373 & 0.9333$\pm$0.0850 & 0.2114$\pm$0.3449 & 0.8614$\pm$0.0419 & 0.6183$\pm$0.1908 \\
&Swish     & 0.7535$\pm$0.1234 & 0.8690$\pm$0.1899 & 0.7767$\pm$0.2051 & 0.6743$\pm$0.3449 & 0.8096$\pm$0.1741 & \textbf{0.8374$\pm$0.0936} \\
&LeakyReLU & 0.7935$\pm$0.0428 & 0.8250$\pm$0.0540 & \textbf{0.9400$\pm$0.0528}
 & 0.2914$\pm$0.2664 & 0.8761$\pm$0.0238 & 0.6931$\pm$0.2170 \\
&ELU       & 0.7913$\pm$0.0610 & \textbf{0.9128$\pm$0.0567}
 & 0.8267$\pm$0.0620 & \textbf{0.7143$\pm$0.2138}
 & 0.8652$\pm$0.0423 & 0.8214$\pm$0.0947 \\
\midrule
HouseVotes84
  & 3C-EA   & $\mathbf{0.9407 \pm 0.0191}^{**\dagger\dagger\ddagger\ddagger\S}$ & $\mathbf{0.9027 \pm 0.0469}^{**\dagger\dagger\ddagger\ddagger\S\S}$ & $0.9503 \pm 0.0476^{*\dagger\ddagger}$ & $\mathbf{0.9348 \pm 0.0352}^{*\dagger\dagger\ddagger\ddagger\S\S}$ & $\mathbf{0.9241 \pm 0.0240}^{**\dagger\dagger\ddagger\ddagger}$ & $\mathbf{0.9858 \pm 0.0088}^{**\dagger\dagger\ddagger\ddagger}$\\
&ReLU      & 0.9126$\pm$0.0220 & 0.8654$\pm$0.0506 & 0.9188$\pm$0.0684 & 0.9089$\pm$0.0352 & 0.8882$\pm$0.0287 & 0.9768$\pm$0.0120 \\
&Swish     & 0.9053$\pm$0.0226 & 0.8485$\pm$0.0492 & 0.9212$\pm$0.0653 & 0.8956$\pm$0.0352 & 0.8805$\pm$0.0277 & 0.9732$\pm$0.0129 \\
&LeakyReLU & 0.9034$\pm$0.0163 & 0.8510$\pm$0.0289 & 0.9067$\pm$0.0599 & 0.9015$\pm$0.0260 & 0.8762$\pm$0.0240 & 0.9744$\pm$0.0117 \\
&ELU       & 0.9255$\pm$0.0276 & 0.8631$\pm$0.0550 & \textbf{0.9612$\pm$0.0398} & 0.9037$\pm$0.0448 & 0.9080$\pm$0.0323 & 0.9792$\pm$0.0138 \\
\midrule
Credit Approval
  & 3C-EA & \textbf{0.8455$\pm$0.0191} & \textbf{0.8096$\pm$0.0321}& $0.8536 \pm 0.0254^{*\dagger}$ & \textbf{0.8420$\pm$0.0363} & $\textbf{0.8366$\pm$0.0180}^{\dagger}$ & $\textbf{0.9092$\pm$0.0158}^{\dagger\dagger}$ \\
& ReLU & $0.8367 \pm 0.0226$ & $0.8086 \pm 0.0492$ & $0.8339 \pm 0.0437$ & $0.8390 \pm 0.0363$ & $0.8189 \pm 0.0210$ & $0.9027 \pm 0.0143$ \\
& Swish & $0.8304 \pm 0.0249$ & $0.8067 \pm 0.0544$ & $0.8208 \pm 0.0446$ & $0.8381 \pm 0.0363$ & $0.8110 \pm 0.0212$ & $0.8960 \pm 0.0136$ \\
& LeakyReLU & $0.8319 \pm 0.0251$ & $0.7997 \pm 0.0417$ & $0.8317 \pm 0.0293$ & $0.8320 \pm 0.0500$ & $0.8143 \pm 0.0230$ & $0.9021 \pm 0.0144$ \\
& ELU & $0.8425 \pm 0.0217$ & $0.7985 \pm 0.0312$ & $\mathbf{0.8628 \pm 0.0241}$ & $0.8264 \pm 0.0329$ & $0.8290 \pm 0.0218$ & $0.9043 \pm 0.0168$ \\
\midrule
Mammographic & 3C-EA & $\mathbf{0.8118 \pm 0.0159}^{**\dagger\dagger\ddagger\S\S}$ & $\mathbf{0.7798 \pm 0.0233}^{\dagger\dagger\S\S}$ & $\mathbf{0.8510 \pm 0.0285}^{\dagger}$ & $\mathbf{0.7753 \pm 0.0323}^{\S\S}$ & $\mathbf{0.8133 \pm 0.0157}^{**\dagger\dagger\ddagger\S\S}$ & $\mathbf{0.8854 \pm 0.0084}^{*\dagger\dagger\S\S}$ \\
& ReLU & $0.7971 \pm 0.0135$ & $0.7652 \pm 0.0286$ & $0.8400 \pm 0.0469$ & $0.7572 \pm 0.0323$ & $0.7993 \pm 0.0150$ & $0.8789 \pm 0.0119$ \\
& Swish & $0.7940 \pm 0.0131$ & $0.7637 \pm 0.0210$ & $0.8315 \pm 0.0399$ & $0.7591 \pm 0.0323$ & $0.7952 \pm 0.0149$ & $0.8706 \pm 0.0126$ \\
& LeakyReLU & $0.8036 \pm 0.0152$ & $0.7740 \pm 0.0345$ & $0.8430 \pm 0.0478$ & $0.7670 \pm 0.0553$ & $0.8051 \pm 0.0149$ & $0.8802 \pm 0.0100$ \\
& ELU & $0.7945 \pm 0.0113$ & $0.7562 \pm 0.0220$ & $0.8490 \pm 0.0335$ & $0.7437 \pm 0.0355$ & $0.7991 \pm 0.0116$ & $0.8606 \pm 0.0115$ \\
\midrule
Cylinder Bands & 3C-EA & $0.6541 \pm 0.0240$ & $0.6812 \pm 0.0186 ^{**\dagger\ddagger\ddagger}$ & $0.7148 \pm 0.0458$ & $0.5487 \pm 0.0418 ^{**\dagger\dagger\ddagger\ddagger}$ & $0.6969 \pm 0.0255$ & $0.6940 \pm 0.0242 ^{**\ddagger}$ \\
& ReLU & $0.6348 \pm 0.0477$ & $0.6294 \pm 0.0424$ & \textbf{0.9129$\pm$0.0830} & $0.2600 \pm 0.0418$ & \textbf{0.7414$\pm$0.0294} & $0.6433 \pm 0.0967$ \\
& Swish & $0.6478 \pm 0.0345$ & $0.6571 \pm 0.0376$ & $0.8316 \pm 0.1035$ & $0.4000 \pm 0.0418$ & $0.7290 \pm 0.0291$ & $0.6622 \pm 0.0718$ \\
& LeakyReLU & $0.6415 \pm 0.0435$ & $0.6433 \pm 0.0378$ & $0.8632 \pm 0.1008$ & $0.3426 \pm 0.1638$ & $0.7329 \pm 0.0373$ & $0.6593 \pm 0.0677$ \\
& ELU & $\mathbf{0.6637 \pm 0.0308}$ & $\mathbf{0.6921 \pm 0.0245}$ & $0.7465 \pm 0.0404$ & $\mathbf{0.5522 \pm 0.0426}$ & $0.7179 \pm 0.0283$ & $\mathbf{0.7130 \pm 0.0271}$ \\
\midrule
Heart Disease & 3C-EA & $\mathbf{0.9198 \pm 0.0482}^{**\dagger\dagger\ddagger\ddagger\S\S}$ & $\mathbf{0.9218 \pm 0.0587}^{**\dagger\dagger\ddagger\ddagger\S\S}$ & $\mathbf{0.9242 \pm 0.0374}^{**\dagger\dagger\ddagger\ddagger\S\S}$ & $\mathbf{0.9152 \pm 0.0681}^{**\dagger\dagger\ddagger\ddagger\S\S}$ & $\mathbf{0.9225 \pm 0.0451}^{**\dagger\dagger\ddagger\ddagger\S\S}$ & $\mathbf{0.9636 \pm 0.0294}^{**\dagger\dagger\ddagger\ddagger\S\S}$ \\
& ReLU & $0.8246 \pm 0.0164$ & $0.8246 \pm 0.0390$ & $0.8415 \pm 0.0622$ & $0.8068 \pm 0.0681$ & $0.8302 \pm 0.0207$ & $0.9112 \pm 0.0103$ \\
& Swish & $0.8041 \pm 0.0615$ & $0.7930 \pm 0.0633$ & $0.8560 \pm 0.0514$ & $0.7496 \pm 0.0681$ & $0.8196 \pm 0.0331$ & $0.8853 \pm 0.1206$ \\
& LeakyReLU & $0.8281 \pm 0.0203$ & $0.8216 \pm 0.0308$ & $0.8518 \pm 0.0486$ & $0.8032 \pm 0.0465$ & $0.8350 \pm 0.0217$ & $0.9167 \pm 0.0116$ \\
& ELU & $0.8353 \pm 0.0184$ & $0.8192 \pm 0.0244$ & $0.8720 \pm 0.0294$ & $0.7968 \pm 0.0346$ & $0.8443 \pm 0.0176$ & $0.9245 \pm 0.0219$ \\
\midrule
Adult & 3C-EA & $\mathbf{0.8241 \pm 0.0055}^{**\dagger\dagger\ddagger\ddagger\S\S}$ & $\mathbf{0.6662 \pm 0.0094}^{**\ddagger\ddagger}$ & $\mathbf{0.5408 \pm 0.0314}^{**\dagger\dagger\ddagger\ddagger\S\S}$ & $0.9141 \pm 0.0047$ & $\mathbf{0.5966 \pm 0.0221}^{**\dagger\dagger\ddagger\ddagger\S\S}$ & $\mathbf{0.8682 \pm 0.0070}^{**\dagger\dagger\ddagger\ddagger\S\S}$ \\
& ReLU & $0.7596 \pm 0.0023$ & $0.2445 \pm 0.2682$ & $0.0101 \pm 0.0213$ & $\mathbf{0.9975 \pm 0.0047}$ & $0.0185 \pm 0.0383$ & $0.5664 \pm 0.0598$ \\
& Swish & $0.7922 \pm 0.0073$ & $0.6326 \pm 0.1289$ & $0.3080 \pm 0.0835$ & $0.9458 \pm 0.0047$ & $0.4077 \pm 0.0966$ & $0.7990 \pm 0.0383$ \\
& LeakyReLU & $0.7654 \pm 0.0132$ & $0.3671 \pm 0.3678$ & $0.0543 \pm 0.1159$ & $0.9911 \pm 0.0198$ & $0.0764 \pm 0.1483$ & $0.6150 \pm 0.0994$ \\
& ELU & $0.7990 \pm 0.0096$ & $0.6589 \pm 0.0180$ & $0.3434 \pm 0.0697$ & $0.9436 \pm 0.0111$ & $0.4476 \pm 0.0587$ & $0.8183 \pm 0.0287$ \\
\midrule
\end{tabular}
\end{table*}

We start our empirical examination on datasets that naturally contain missing values, matching real-world techniques for collecting data where incompleteness occurs naturally rather than through controlled simulation. For these datasets, no artificial missingness is introduced. The models function directly on the original incomplete data distributions. All datasets are divided into training, validation, and test sets utilizing fixed splits that are uniformly applied across all methods to guarantee fair comparison.
Table \ref{tab:Incomplete} summarizes the statistical results for all real world incomplete datasets. Performance is assessed by six standard metrics: accuracy, precision, sensitivity (recall), specificity, F1-score, and area under the ROC curve (AUC).

% The experiments show a clear advantage for 3C-EA, which achieves the highest test accuracy on six out of seven datasets (Table \ref{tab:Incomplete})
The results in Table~\ref{tab:Incomplete} show that 3C-EA is competitive and achieves the highest test accuracy on six out of seven datasets. On Heart Disease, 3C-EA has the best test accuracy (0.9198) and F1-score (0.9225), which is a 9 percentage point improvement over ReLU.
% On HouseVotes84 and Credit Approval, 3C-EA clearly outperforms baselines across all metrics, with HouseVotes84 achieving accuracy 0.9407 and AUC 0.9858.
On HouseVotes84 and Credit Approval, 3C-EA shows strong overall performance, including the highest test accuracy on both datasets; on HouseVotes84, it reaches 0.9407 accuracy and 0.9858 AUC.
On Cylinder Bands, where the baseline methods are challenged by 38\% missing data, 3C-EA shows a large improvement in specificity (0.5487 vs. 0.2600 for ReLU), indicating a better performance in the presence of poor-quality data. The Adult dataset serves as a severe stress test because it contains many samples (48,842), is naturally incomplete, and is extremely imbalanced. 
In this scenario, 3C-EA has the best overall accuracy (0.8241) and a substantially higher recall (0.5408), suggesting that it is not simply collapsing to majority-class predictions. This suggests  that modifying activations on value, missingness, and confidence can help the network to more effectively manage real-world incomplete tabular data by differentiating between observed and imputed inputs.
% \subsubsection{Genetic Programming Parameters}
% The evolutionary search was configured using the table \ref{tab:hyperparams}.

\subsection{Results on Complete Datasets}

We evaluate our method (3C-EA) against missingness by adding 20\% missingness to originally complete datasets using the MNAR, MCAR, and MAR methods. In order to prevent degenerate sparsity, this fixed rate allows for direct comparison among techniques. All approaches use the same preprocessing, divisions, architectures, and training protocols.

\subsubsection{MNAR Missingness}

\begin{table*}[htbp]
\caption{MNAR Imputed datasets  results. }
\label{tab:mnar}
\renewcommand{\arraystretch}{0.70}
\centering
\scriptsize
\setlength{\tabcolsep}{4pt}
\setlength{\abovecaptionskip}{2pt}
\setlength{\belowcaptionskip}{0pt}
\begin{tabular}{llcccccc}
\toprule
Dataset & Method & TestAcc & Prec & Rec & Spec & F1 & AUC \\
\midrule
Sonar & 3C-EA & $\mathbf{0.7619 \pm 0.0447}$ & $\mathbf{0.7591 \pm 0.0536}^{*}$ & $\mathbf{0.8106 \pm 0.0722}^{\dagger\S\S}$ & $0.7160 \pm 0.0674^{*}$ & $\mathbf{0.7791 \pm 0.0423}^{{\dagger\S\S}}$ & $\mathbf{0.8395 \pm 0.0304}^{*\S\S}$\\
     & ReLU    & 0.7467$\pm$0.0499 & 0.7382$\pm$0.0448 & 0.8073$\pm$0.1215 & 0.6800$\pm$0.0674 & 0.7655$\pm$0.0616 & 0.8190$\pm$0.0481 \\
     & Swish   & 0.7400$\pm$0.0495 & 0.7546$\pm$0.0504 & 0.7618$\pm$0.1422 & 0.7160$\pm$0.0674 & 0.7479$\pm$0.0724 & 0.8214$\pm$0.0360 \\
&LeakyReLU & 0.7393$\pm$0.0352 & 0.7567$\pm$0.0453 & 0.8055$\pm$0.0905 & 0.7080$\pm$0.0868 & 0.7761$\pm$0.0410 & 0.8368$\pm$0.0395 \\
&ELU      & 0.7362$\pm$0.0409 & 0.7586$\pm$0.0670 & 0.7418$\pm$0.0526 & \textbf{0.7300$\pm$0.0990} & 0.7469$\pm$0.0335 & 0.7994$\pm$0.0422 \\
\midrule
Glass &3C-EA   & $\mathbf{0.5247\pm 0.0784}^{\dagger}$ & $\mathbf{0.3381 \pm 0.0972}$ & $\mathbf{0.3773 \pm 0.0812}$ & ${0.8874 \pm 0.0181}$ & $\mathbf{0.3390 \pm 0.0868}$ & $\mathbf{0.7729 \pm 0.0745}$
 \\
&ReLU      & 0.4065$\pm$0.1059 & 0.2093$\pm$0.0938 & 0.2839$\pm$0.0681 & 0.8874$\pm$0.0181 & 0.2124$\pm$0.0759 & 0.6475$\pm$0.0832 \\
&Swish     & 0.4400$\pm$0.0627 & 0.2535$\pm$0.0829 & 0.3152$\pm$0.0383 & 0.8874$\pm$0.0181 & 0.2540$\pm$0.0531 & 0.7300$\pm$0.0603 \\
&LeakyReLU & 0.4521$\pm$0.0535 & 0.2323$\pm$0.1057 & 0.2935$\pm$0.0621 & 0.8874$\pm$0.0181 & 0.2361$\pm$0.0796 & 0.6486$\pm$0.0872 \\
&ELU       & 0.4977$\pm$0.0724 & 0.3313$\pm$0.0933 & 0.3537$\pm$0.0645 & 0.8874$\pm$0.0181 & 0.3112$\pm$0.0680 & 0.7025$\pm$0.0752 \\
\midrule
WDBC & 3C-EA   & $\mathbf{0.9312\pm0.0273}$ & $\mathbf{0.9281\pm0.0428}$ & $\mathbf{0.8914\pm0.0621}^{*}$ & $\mathbf{0.9591\pm0.0721}$ & $\mathbf{0.9045\pm0.0405}$ & $0.9733\pm0.0241$ \\
&ReLU      & 0.9221$\pm$0.0205 & 0.9262$\pm$0.0441 & 0.8610$\pm$0.0550 & 0.9578$\pm$0.0265 & 0.8903$\pm$0.0290 & \textbf{0.9835$\pm$0.0069} \\
&Swish     & 0.9196$\pm$0.0156 & 0.8997$\pm$0.0503 & 0.8857$\pm$0.0442 & 0.9394$\pm$0.0265 & 0.8906$\pm$0.0194 & 0.9833$\pm$0.0052 \\
&LeakyReLU & 0.9232$\pm$0.0157 & 0.9184$\pm$0.0414 & 0.8724$\pm$0.0542 & 0.9528$\pm$0.0269 & 0.8928$\pm$0.0239 & 0.9814$\pm$0.0129 \\
&ELU       & 0.9263$\pm$0.0233 & 0.9142$\pm$0.0557 & 0.8876$\pm$0.0372 & 0.9489$\pm$0.0364 & 0.8992$\pm$0.0295 & 0.9819$\pm$0.0129 \\
\midrule
PIMA &3C-EA    & $\mathbf{0.7042\pm0.0209}$ & $0.6043\pm0.0497$ & $\mathbf{0.4121\pm0.0763}^{**\dagger\dagger}$ & $0.8574\pm0.0350$ & ${0.4856\pm0.0642}^{*\dagger\dagger}$ & $\mathbf{0.7306\pm0.0376}^{\dagger\dagger}$ \\
&ReLU       & $0.6956\pm0.0276$ & $0.6310\pm0.1654$ & $0.2604\pm0.1395$ & $\mathbf{0.9240\pm0.0350}$ & $0.3485\pm0.1477$ & $0.7073\pm0.0643$ \\
&Swish      & 0.7005$\pm$0.0213 & 0.6260$\pm$0.0718 & 0.3758$\pm$0.1482 & 0.8709$\pm$0.0350 & 0.4457$\pm$0.1148 & 0.7343$\pm$0.0234 \\
&LeakyReLU  & 0.7016$\pm$0.0281 & \textbf{0.6480$\pm$0.0757} & 0.3192$\pm$0.1103 & 0.9022$\pm$0.0639 & 0.4135$\pm$0.0965 & 0.7305$\pm$0.0314 \\
&ELU        & 0.7022$\pm$0.0152 & 0.6294$\pm$0.0469 & 0.4113$\pm$0.0562 & 0.8701$\pm$0.0346 & $\mathbf{0.4940 \pm 0.0393}$ & 0.7274$\pm$0.0189 \\
\midrule
Mushroom & 3C-EA & $\mathbf{0.9850 \pm 0.0097}^{**\dagger\dagger\ddagger\ddagger\S\S}$ & $\mathbf{0.9831 \pm 0.0092}^{**\dagger\dagger\ddagger\ddagger\S\S}$ & $\mathbf{0.9858 \pm 0.0118}^{**\dagger\dagger\ddagger\ddagger\S\S}$ & $\mathbf{0.9850 \pm 0.0098}^{**\dagger\dagger\ddagger\ddagger\S\S}$ & $\mathbf{0.9844 \pm 0.0101}^{**\dagger\dagger\ddagger\ddagger\S\S}$ & $\mathbf{0.9973 \pm 0.0038}^{**\dagger\dagger\ddagger\ddagger\S\S}$ \\
& ReLU & $0.9426 \pm 0.0284$ & $0.9637 \pm 0.0084$ & $0.9153 \pm 0.0580$ & $0.9417 \pm 0.0294$ & $0.9380 \pm 0.0326$ & $0.9807 \pm 0.0174$ \\
& Swish & $0.9086 \pm 0.0192$ & $0.9383 \pm 0.0136$ & $0.8674 \pm 0.0386$ & $0.9071 \pm 0.0198$ & $0.9010 \pm 0.0224$ & $0.9557 \pm 0.0239$ \\
& LeakyReLU & $0.9399 \pm 0.0276$ & $0.9571 \pm 0.0131$ & $0.9162 \pm 0.0517$ & $0.9391 \pm 0.0284$ & $0.9356 \pm 0.0311$ & $0.9797 \pm 0.0187$ \\
& ELU & $0.9666 \pm 0.0201$ & $0.9696 \pm 0.0143$ & $0.9606 \pm 0.0292$ & $0.9664 \pm 0.0205$ & $0.9650 \pm 0.0217$ & $0.9933 \pm 0.0089$ \\
\bottomrule
\end{tabular}
\end{table*}

MNAR is the most difficult case, in which the missingness is related to the values themselves. The results are shown in Table \ref{tab:mnar}. In the six-class imbalanced, Glass dataset, 3C-EA shows an accuracy of 52.5\%, surpassing the baseline range of 40.7\% to 49.8\% (an improvement of 2.7\% to 11.8\%). The F1-score rises to 0.339 from a range of 0.210–0.311, while the AUC improves to 0.773 from a range of 0.648–0.730. The reason for this is ChannelProp: in MNAR, extreme values are more vulnerable to being missing, and the evolved functions of 3C-EA explicitly condition on distributed confidence scores to modify responses accordingly. On the PIMA dataset, in addition to that, our method shows the highest values on accuracy; it also has the highest recall of 0.412, surpassing ReLU (0.260), Swish (0.376), and LeakyReLU (0.319), while ELU is roughly comparable at 0.411. Regarding the F1-score, 3C-EA (0.486) surpassed ReLU, Swish, and LeakyReLU, however ELU had somewhat superior performance (0.494). The increase in recall demonstrates that reliability-aware activations prevent majority class collapse in uncertain conditions. Also, 3C-EA displays the best accuracy (93.1\%) and F1-score (0.905), which shows that it works well on WDBC, a clinical breast cancer dataset.
In Sonar with 60 features, 3C-EA has the highest accuracy (76.2\%), F1-score (0.779), and AUC (0.840) among the methods, showing that ChannelProp effectively scales in high-dimensional datasets. Additionally, in the Mushroom dataset with 8,124 samples, it has the highest accuracy of 98.5\% with the lowest variance of $\pm$0.0097 when compared to baselines of $\pm$0.019--0.028, indicating stable large-scale optimization.

\subsubsection{MCAR Missingness}

\begin{table*}[htbp]
\caption{MCAR Imputed datasets results. }
\label{tab:mcar}
\renewcommand{\arraystretch}{0.70}
\centering
\scriptsize
\setlength{\tabcolsep}{4pt}
\setlength{\abovecaptionskip}{2pt}
\setlength{\belowcaptionskip}{0pt}

\begin{tabular}{llcccccc}
\toprule
Dataset & Method & TestAcc & Prec & Rec & Spec & F1 & AUC \\
\midrule
Sonar & 3C-EA & $0.7413 \pm 0.0626$ & $0.7436 \pm 0.0573^{*}$ & $0.7803 \pm 0.1214^{\dagger\S\S}$ & $0.6983 \pm 0.1099$ & $0.7561 \pm 0.0717^{\dagger}$ & $0.8084 \pm 0.0626$ \\
& ReLU & $0.7437 \pm 0.0635$ & $0.7197 \pm 0.0590$ & $\mathbf{0.8561 \pm 0.1144}$ & $0.6200 \pm 0.1099$ & $\mathbf{0.7756 \pm 0.0633}$ & $0.8208 \pm 0.0799$ \\
& Swish & $0.7405 \pm 0.0666$ & $0.7187 \pm 0.0531$ & $0.8379 \pm 0.1310$ & $0.6333 \pm 0.1099$ & $0.7680 \pm 0.0701$ & $0.8205 \pm 0.0505$ \\
& LeakyReLU & $0.7468 \pm 0.0613$ & $0.7316 \pm 0.0645$ & $0.8409 \pm 0.1455$ & $0.6433 \pm 0.1377$ & $0.7714 \pm 0.0739$ & $\mathbf{0.8300 \pm 0.0397}$ \\
& ELU & $\mathbf{0.7556 \pm 0.0529}$ & $\mathbf{0.7625 \pm 0.0593}$ & $0.7833 \pm 0.0884$ & $\mathbf{0.7250 \pm 0.0920}$ & $0.7691 \pm 0.0550$ & $0.8155 \pm 0.0544$ \\
\bottomrule
Glass & 3C-EA & $\mathbf{0.5318 \pm 0.1038}^{**\dagger\dagger\ddagger}$ & $\mathbf{0.3922 \pm 0.1411}^{**\dagger\dagger\ddagger}$ & $0.4023 \pm 0.0939$ & $\mathbf{0.8896 \pm 0.0238}$ & $\mathbf{0.3668 \pm 0.1165}^{**\dagger\dagger\ddagger\ddagger}$ & $\mathbf{0.7677 \pm 0.1037}^{*}$ \\
& ReLU & $0.4550 \pm 0.0906$ & $0.2525 \pm 0.1043$ & $0.2994 \pm 0.0689$ & $\mathbf{0.8896 \pm 0.0238}$ & $0.2423 \pm 0.0811$ & $0.7068 \pm 0.0782$ \\
& Swish & $0.4558 \pm 0.0873$ & $0.2554 \pm 0.0964$ & $0.3198 \pm 0.0711$ & $\mathbf{0.8896 \pm 0.0238}$ & $0.2534 \pm 0.0813$ & $0.7575 \pm 0.0953$ \\
& LeakyReLU & $0.4721 \pm 0.0682$ & $0.2998 \pm 0.0969$ & $0.3246 \pm 0.0620$ & $\mathbf{0.8896 \pm 0.0238}$ & $0.2758 \pm 0.0682$ & $0.7462 \pm 0.0855$ \\
& ELU & $0.5023 \pm 0.1043$ & $0.3550 \pm 0.1318$ & $\mathbf{0.3801 \pm 0.1054}$ & $\mathbf{0.8896 \pm 0.0238}$ & $0.3361 \pm 0.1143$ & $0.7475 \pm 0.0896$ \\
\midrule
WDBC & 3C-EA & $\mathbf{0.9415 \pm 0.0154}^{**\dagger\dagger\ddagger\ddagger\S}$ & $0.9383 \pm 0.0367^{\dagger\dagger}$ & $\mathbf{0.9032 \pm 0.0464}^{**\dagger\ddagger\ddagger\S\S}$ & $0.9639 \pm 0.0228$ & $\mathbf{0.9189 \pm 0.0233}^{**\dagger\dagger\ddagger\ddagger\S}$ & $\mathbf{0.9832 \pm 0.0102}$ \\
& ReLU & $0.9246 \pm 0.0163$ & $0.9399 \pm 0.0461$ & $0.8532 \pm 0.0435$ & $0.9662 \pm 0.0228$ & $0.8928 \pm 0.0232$ & $0.9821 \pm 0.0079$ \\
& Swish & $0.9164 \pm 0.0249$ & $0.9029 \pm 0.0684$ & $0.8770 \pm 0.0587$ & $0.9394 \pm 0.0228$ & $0.8859 \pm 0.0300$ & $0.9822 \pm 0.0056$ \\
& LeakyReLU & $0.9246 \pm 0.0208$ & $\mathbf{0.9478 \pm 0.0546}$ & $0.8468 \pm 0.0545$ & $\mathbf{0.9699 \pm 0.0342}$ & $0.8919 \pm 0.0303$ & $0.9806 \pm 0.0123$ \\
& ELU & $0.9316 \pm 0.0195$ & $0.9361 \pm 0.0454$ & $0.8770 \pm 0.0384$ & $0.9634 \pm 0.0286$ & $0.9043 \pm 0.0263$ & $0.9831 \pm 0.0109$ \\
\bottomrule
PIMA & 3C-EA & $\mathbf{0.7093 \pm 0.0204}^{*}$ & $0.6102 \pm 0.0457$ & $\mathbf{0.4687 \pm 0.0449}^{**\dagger\dagger\ddagger\ddagger}$ & $0.8325 \pm 0.0426$ & $\mathbf{0.5237 \pm 0.0262}^{**\dagger\dagger\ddagger\ddagger}$ & $\mathbf{0.7437 \pm 0.0194}^{**\dagger\dagger\ddagger\ddagger\S}$ \\

& ReLU & $0.6958 \pm 0.0249$ & $0.6249 \pm 0.1570$ & $0.2823 \pm 0.1254$ & $\mathbf{0.9129 \pm 0.0426}$ & $0.3714 \pm 0.1382$ & $0.6919 \pm 0.0543$ \\

& Swish & $0.7005 \pm 0.0230$ & $0.6285 \pm 0.0924$ & $0.3819 \pm 0.1235$ & $0.8677 \pm 0.0426$ & $0.4552 \pm 0.0929$ & $0.7261 \pm 0.0205$ \\

& LeakyReLU & $0.7075 \pm 0.0268$ & $\mathbf{0.6547 \pm 0.0679}$ & $0.3442 \pm 0.0781$ & $0.8982 \pm 0.0554$ & $0.4425 \pm 0.0700$ & $0.7216 \pm 0.0355$ \\

& ELU & $0.7034 \pm 0.0323$ & $0.6033 \pm 0.0673$ & $0.4370 \pm 0.0862$ & $0.8432 \pm 0.0611$ & $0.4998 \pm 0.0633$ & $0.7292 \pm 0.0334$ \\
\midrule
Mushroom & 3C-EA & $\mathbf{0.9843 \pm 0.0034}^{**\dagger\dagger\ddagger\ddagger\S\S}$ & $\mathbf{0.9828 \pm 0.0047}^{**\dagger\dagger\ddagger\ddagger\S\S}$ & $\mathbf{0.9847 \pm 0.0041}^{**\dagger\dagger\ddagger\ddagger\S\S}$ & $\mathbf{0.9843 \pm 0.0031}^{**\dagger\dagger\ddagger\ddagger\S\S}$ & $\mathbf{0.9837 \pm 0.0035}^{**\dagger\dagger\ddagger\ddagger\S\S}$ & $\mathbf{0.9957 \pm 0.0039}^{**\dagger\dagger\ddagger\ddagger\S\S}$ \\
& ReLU & $0.9379 \pm 0.0311$ & $0.9537 \pm 0.0175$ & $0.9165 \pm 0.0491$ & $0.9377 \pm 0.0295$ & $0.9343 \pm 0.0320$ & $0.9831 \pm 0.0215$ \\ & Swish & $0.9073 \pm 0.0212$ & $0.9320 \pm 0.0181$ & $0.8713 \pm 0.0375$ & $0.9060 \pm 0.0217$ & $0.9002 \pm 0.0239$ & $0.9573 \pm 0.0327$ \\
& LeakyReLU & $0.9385 \pm 0.0283$ & $0.9558 \pm 0.0137$ & $0.9147 \pm 0.0524$ & $0.9378 \pm 0.0291$ & $0.9391 \pm 0.0473$ & $0.9791 \pm 0.0219$ \\
& ELU & $0.9671 \pm 0.0207$ & $0.9682 \pm 0.0149$ & $0.9591 \pm 0.0298$ & $0.9649 \pm 0.0211$ & $0.9638 \pm 0.0543$ & $0.9921 \pm 0.0095$ \\
\midrule
\end{tabular}
\end{table*}

\begin{figure}[htbp]
    \captionsetup{font=small}
    \centering
    \includegraphics[width=0.9\columnwidth]{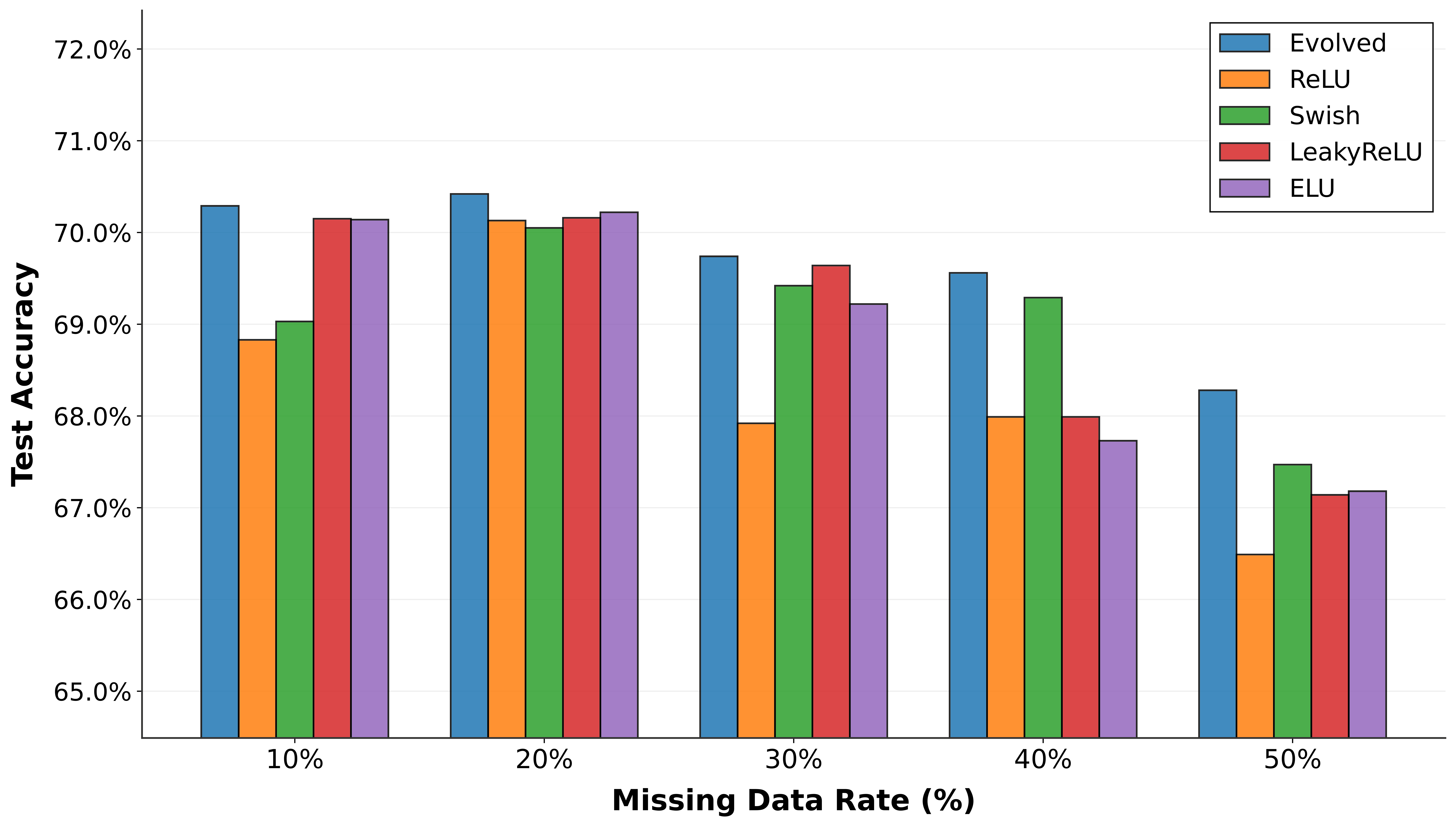}
    \caption{\small Performance comparison across missing data rates (10\%-50\%) on the PIMA dataset.}
    \Description{}
    \label{fig:performance}
\end{figure}

Results related to MCAR missingness are presented in Table \ref{tab:mcar}. On the Glass dataset, regarding uninformative missingness, 3C-EA achieves the highest accuracy of 53.2\%, surpassing the baselines which range from 45.5\% to 50.2\%. The F1-score improves from 0.242–0.336 to 0.367. This shows that confidence data is useful even when the missingness patterns don't contain any information, GP-evolved activations are able to determine appropriate weights to imputed versus actual values depending on feature-specific reliability.
Our method achieves the highest recall (0.469), F1-score (0.524), and AUC (0.744) on PIMA in comparison to the baseline models. The recall has increased significantly compared to the baselines, confirming that the three-channel activation provides a comprehensive strategy to handle uncertainty, rather than only exploiting patterns of missing data.
Figure \ref{fig:performance} illustrates the effectiveness of our approach on PIMA with missing rates ranging from 10\% to 50\%. The performance gap between 3C-EA and the baselines grows with the missing rate, showing that the reliability information is more beneficial in conditions of poor data quality.

\subsubsection{MAR Missingness}

\begin{table*}[htbp]
\caption{MAR Imputed datasets results. }
\label{tab:mar}
\renewcommand{\arraystretch}{0.70}
\centering
\scriptsize
\setlength{\tabcolsep}{4pt}
\setlength{\abovecaptionskip}{2pt}
\setlength{\belowcaptionskip}{0pt}
\begin{tabular}{llcccccc}
\toprule
Dataset & Method & TestAcc & Prec & Rec & Spec & F1 & AUC \\
\midrule
Sonar &  3C-EA & $0.7691 \pm 0.0683$ & $0.7611 \pm 0.0747$ & $0.8030 \pm 0.0956$ & $0.7067 \pm 0.1574$ & $0.7760 \pm 0.0565$ & $0.8368 \pm 0.0592$ \\
& ReLU & $\mathbf{0.7802 \pm 0.0661}$ & $0.7759 \pm 0.0668$ & $0.8273 \pm 0.1304$ & $0.7283 \pm 0.1574$ & \textbf{0.7936$\pm$0.0751} & $0.8467 \pm 0.0445$ \\
& Swish & $0.7778 \pm 0.0489$ & $0.7724 \pm 0.0491$ & $\mathbf{0.8303 \pm 0.1378}$ & $0.7200 \pm 0.1574$ & $0.7913 \pm 0.0655$ & $\mathbf{0.8524 \pm 0.0298}$ \\
& LeakyReLU & $0.7595 \pm 0.0613$ & $0.7556 \pm 0.0754$ & $0.8242 \pm 0.1166$ & $0.6883 \pm 0.1453$ & $0.7803 \pm 0.0587$ & $0.8415 \pm 0.0520$ \\
& ELU & $0.7706 \pm 0.0578$ & $\mathbf{0.7820 \pm 0.0632}$ & $0.7864 \pm 0.0848$ & $\mathbf{0.7533 \pm 0.0912}$ & $0.7812 \pm 0.0585$ & $0.8286 \pm 0.0548$ \\
\midrule
Glass & 3C-EA & $\mathbf{0.5803 \pm 0.1060}^{**\dagger\dagger\ddagger\ddagger}$ & $\mathbf{0.3880 \pm 0.1415}^{**\dagger\dagger\ddagger\ddagger}$ & $0.3952 \pm 0.0931^{**\dagger\dagger\ddagger\ddagger}$ & $\mathbf{0.8976 \pm 0.0264}$ & $0.3721 \pm 0.1118^{**\dagger\dagger\ddagger\ddagger}$ & $\mathbf{0.7778 \pm 0.1080}^{**\dagger\dagger\ddagger\ddagger\S\S}$ \\
& ReLU & $0.4558 \pm 0.1184$ & $0.2655 \pm 0.1061$ & $0.2988 \pm 0.0638$ & $\mathbf{0.8976 \pm 0.0264}$ & $0.2495 \pm 0.0836$ & $0.6645 \pm 0.0835$ \\
& Swish & $0.4736 \pm 0.0912$ & $0.2923 \pm 0.0751$ & $0.3277 \pm 0.0602$ & $\mathbf{0.8976 \pm 0.0264}$ & $0.2751 \pm 0.0651$ & $0.7366 \pm 0.0751$ \\
& LeakyReLU & $0.4837 \pm 0.1018$ & $0.2726 \pm 0.0933$ & $0.3173 \pm 0.0574$ & $\mathbf{0.8976 \pm 0.0264}$ & $0.2676 \pm 0.0733$ & $0.7062 \pm 0.0517$ \\
& ELU & $0.5787 \pm 0.0753$ & $0.3796 \pm 0.1090$ & $\mathbf{0.4081 \pm 0.0559}$ & $\mathbf{0.8976 \pm 0.0264}$ & $\mathbf{0.3735 \pm 0.0772}$ & $0.7542 \pm 0.0574$ \\
\midrule
WDBC & 3C-EA & $\mathbf{0.9354 \pm 0.0212}$ & $\mathbf{0.9176 \pm 0.0381}^{\S}$ & $0.9127 \pm 0.0423^{*}$ & $\mathbf{0.9546 \pm 0.0254}$ & $\mathbf{0.9123 \pm 0.0288}$ & $\mathbf{0.9896 \pm 0.0169}$ \\
& ReLU & $0.9272 \pm 0.0211$ & $0.9153 \pm 0.0426$ & $0.8881 \pm 0.0587$ & $0.9500 \pm 0.0254$ & $0.8994 \pm 0.0302$ & $0.9845 \pm 0.0087$ \\
& Swish & $0.9278 \pm 0.0226$ & $0.8929 \pm 0.0514$ & $0.9198 \pm 0.0531$ & $0.9324 \pm 0.0254$ & $0.9039 \pm 0.0285$ & $0.9848 \pm 0.0053$ \\
& LeakyReLU & $0.9287 \pm 0.0185$ & $0.9045 \pm 0.0437$ & $0.9056 \pm 0.0474$ & $0.9421 \pm 0.0317$ & $0.9034 \pm 0.0246$ & $0.9846 \pm 0.0057$ \\
& ELU & $0.9322 \pm 0.0277$ & $0.8882 \pm 0.0586$ & $\mathbf{0.9405 \pm 0.0397}$ & $0.9273 \pm 0.0478$ & $0.9117 \pm 0.0322$ & $0.9796 \pm 0.0170$ \\
\midrule
PIMA & 3C-EA & $\mathbf{0.7021 \pm 0.0203}^{\S}$ & $0.5833 \pm 0.0419^{\S}$ & $\mathbf{0.4928 \pm 0.0669}^{**\ddagger\ddagger}$ & $0.8119 \pm 0.0458$ & $\mathbf{0.5304 \pm 0.0395}^{**\dagger\ddagger\ddagger}$ & $\mathbf{0.7428 \pm 0.0209}$ \\
& ReLU & $0.6969 \pm 0.0243$ & $\mathbf{0.6283 \pm 0.0718}$ & $0.3094 \pm 0.1252$ & $\mathbf{0.9002 \pm 0.0458}$ & $0.3964 \pm 0.1240$ & $0.7200 \pm 0.0607$ \\
& Swish & $0.6909 \pm 0.0200$ & $0.5769 \pm 0.0578$ & $0.4619 \pm 0.1279$ & $0.8111 \pm 0.0458$ & $0.4970 \pm 0.0749$ & $0.7367 \pm 0.0199$ \\
& LeakyReLU & $0.6930 \pm 0.0238$ & $0.6179 \pm 0.0797$ & $0.3072 \pm 0.1118$ & $0.8954 \pm 0.0519$ & $0.3955 \pm 0.1066$ & $0.7371 \pm 0.0326$ \\
& ELU & $0.6888 \pm 0.0221$ & $0.5581 \pm 0.0400$ & $0.4808 \pm 0.0742$ & $0.7980 \pm 0.0437$ & $0.5129 \pm 0.0462$ & $0.7294 \pm 0.0348$ \\
\midrule
Mushroom & 3C-EA & $\mathbf{0.9738 \pm 0.0041}^{**\dagger\dagger\ddagger\ddagger\S\S}$ & $\mathbf{0.9682 \pm 0.0055}^{**\dagger\dagger\ddagger\ddagger\S\S}$ & $\mathbf{0.9777 \pm 0.0049}^{**\dagger\dagger\ddagger\ddagger\S\S}$ & $\mathbf{0.9739 \pm 0.0072}^{**\dagger\dagger\ddagger\ddagger\S\S}$ & $\mathbf{0.9729 \pm 0.0042}^{**\dagger\dagger\ddagger\ddagger\S\S}$ & $\mathbf{0.9955 \pm 0.0013}^{**\dagger\dagger\ddagger\ddagger\S\S}$ \\
& ReLU & $0.9428 \pm 0.0190$ & $0.9451 \pm 0.0065$ & $0.9358 \pm 0.0418$ & $0.9426 \pm 0.0198$ & $0.9399 \pm 0.0221$ & $0.9842 \pm 0.0153$ \\
& Swish & $0.9009 \pm 0.0161$ & $0.9191 \pm 0.0116$ & $0.8712 \pm 0.0324$ & $0.8999 \pm 0.0166$ & $0.8942 \pm 0.0187$ & $0.9563 \pm 0.0185$ \\
& LeakyReLU & $0.9244 \pm 0.0211$ & $0.9292 \pm 0.0123$ & $0.9125 \pm 0.0389$ & $0.9240 \pm 0.0217$ & $0.9204 \pm 0.0236$ & $0.9748 \pm 0.0159$ \\
& ELU & $0.9609 \pm 0.0078$ & $0.9564 \pm 0.0088$ & $0.9628 \pm 0.0094$ & $0.9610 \pm 0.0078$ & $0.9596 \pm 0.0081$ & $0.9932 \pm 0.0027$ \\
\midrule
\end{tabular}
\end{table*}

Table \ref{tab:mar} presents results under MAR missingness, when the missingness is conditional upon observable features. In this context, 3C-EA shows its most significant advantages on more difficult datasets. On Glass, 3C-EA demonstrates optimal performance, achieving 58.0\% accuracy, in contrast to the baseline range of 45.6–57.9\%. While ELU exhibits a slightly better F1-score (0.374 versus 0.372), 3C-EA demonstrates the highest AUC (0.778) in contrast to the range of 0.665–0.754, indicating enhanced probability calibration across all six classes. The superior performance of MAR is credited to ChannelProp's effective representation of the conditional missingness pattern in the network, permitting evolved activations to learn from the features responsible for the missing values. On PIMA, our method achieves the highest accuracy (70.2\%), recall (0.493), F1-score (0.530), and AUC (0.743) compared to all other approaches. The relative recall increases are most significant for ReLU (18.3 points) and LeakyReLU (18.6 points). On WDBC, 3C-EA achieves the highest accuracy (93.5\%), precision (0.918), specificity (0.955), F1-score (0.912), and AUC (0.990), confirming consistent improvements in this medical diagnostic test. While 3C-EA does not dominate every dataset-setting pair, the Wilcoxon analysis shows that its gains are statistically supported in many challenging cases, while on other datasets it remains competitive with strong fixed-activation baselines.

\subsection{Ablation Studies}
To evaluate the contribution of each component in our method, we run ablation experiments on four selected datasets (Glass, Sonar, WDBC, PIMA) under challenging circumstances, including MNAR missingness with a missing rate of 40\% and a population size of 50. The reason for choosing such challenging settings is that MNAR is the most difficult type of missing data~\cite{van2012flexible}, where the missingness is a function of the unobserved data itself, and the missing rate of 40\% is chosen to make the reliability information even more important for prediction. The results in Table~\ref{tab:ablation_results} show that removing any element  often reduces performance, although the effect is dataset dependent, thus demonstrating the contribution  of missingness indicators, confidence propagation, and the joint modeling of both in evolved activation functions.

\begin{table}[thbp]
\centering
\renewcommand{\arraystretch}{0.74}
\setlength{\abovecaptionskip}{2pt}
\setlength{\belowcaptionskip}{0pt}
\caption{Ablation Study Results.}
\label{tab:ablation_results}
\footnotesize
\setlength{\tabcolsep}{3pt}
\begin{tabular}{llccc}
\toprule
Dataset & Variant & TestAcc & F1 & AUC \\
\midrule
Glass & Full $f(x,m,c)$ & $\mathbf{0.435 \pm 0.116}$ & $\mathbf{0.229 \pm 0.072}$ & $\mathbf{0.641 \pm 0.233}$\\
& No Conf. $f(x,m)$ & $0.407 \pm 0.059$ & $0.223 \pm 0.075$ & $0.626 \pm 0.085$ \\
& No Flag $f(x,c)$ & $0.386 \pm 0.085$ & $0.220 \pm 0.063$ & 0.631$ \pm 0.050$ \\
& No ChannelProp & $0.381 \pm 0.054$ & $0.200 \pm 0.066$ & $0.611 \pm 0.069$ \\
\midrule
Sonar & Full $f(x,m,c)$ & $\mathbf{0.724 \pm 0.043}$ & $\mathbf{0.733 \pm 0.061}$ & $0.786 \pm 0.036$ \\
& No Conf. $f(x,m)$ & $0.712 \pm 0.048$ & $0.721 \pm 0.071$ & $0.780 \pm 0.054$ \\
& No Flag $f(x,c)$ & $0.650 \pm 0.088$ & $0.632 \pm 0.222$ & $0.721 \pm 0.080$ \\
& No ChannelProp & $0.686 \pm 0.089$ & $0.700 \pm 0.073$ & $\mathbf{0.796 \pm 0.036}$ \\
\midrule
WDBC& Full $f(x,m,c)$   & \textbf{0.9223$\pm$0.0261} & 0.8866$\pm$0.0527 & 0.9693$\pm$0.0160 \\
&No Conf. $f(x,m)$   & 0.9149$\pm$0.0215 & 0.8828$\pm$0.0346 & 0.9612$\pm$0.0155 \\
&No Flag $f(x,c)$ & 0.9161$\pm$0.0346 & \textbf{0.8963$\pm$0.0417} & \textbf{0.9768$\pm$0.0146} \\
&No ChannelProp         & 0.9123$\pm$0.0299 & 0.8814$\pm$0.0440 & 0.9718$\pm$0.0135 \\
\midrule
PIMA & Full $f(x,m,c)$     & \textbf{0.6922$\pm$0.0301} & \textbf{0.4637$\pm$0.1090} & \textbf{0.7174$\pm$0.0311} \\
&No Conf. $f(x,m)$     & 0.6782$\pm$0.0352 & 0.3980$\pm$0.0980 & 0.7005$\pm$0.0536 \\
&No Flag $f(x,c)$   & 0.6741$\pm$0.0211 & 0.4338$\pm$0.0913 & 0.7060$\pm$0.0455 \\
&No ChannelProp           & 0.6701$\pm$0.0258 & 0.4099$\pm$0.1422 & 0.7074$\pm$0.0340 \\
\bottomrule
\end{tabular}
\end{table}

\subsection{Evolved Activation Function Analysis}

\begin{figure}[htbp]
    \captionsetup{font=small}
    \centering
    \includegraphics[width=0.7\columnwidth]{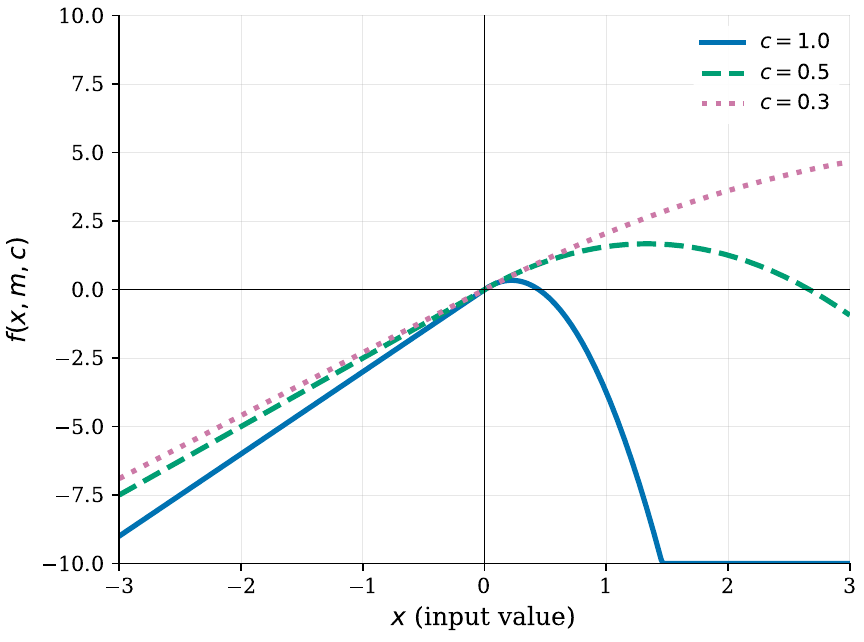}
    \vspace{-2ex}
    \caption{\small Evolved activation function on Heart Disease dataset demonstrating three-channel confidence effect.}
    \Description{}
    \label{fig:missingpima}
\end{figure}

\begin{figure}[htbp]
    \captionsetup{font=small}
    \centering
    \includegraphics[width=0.7\columnwidth]{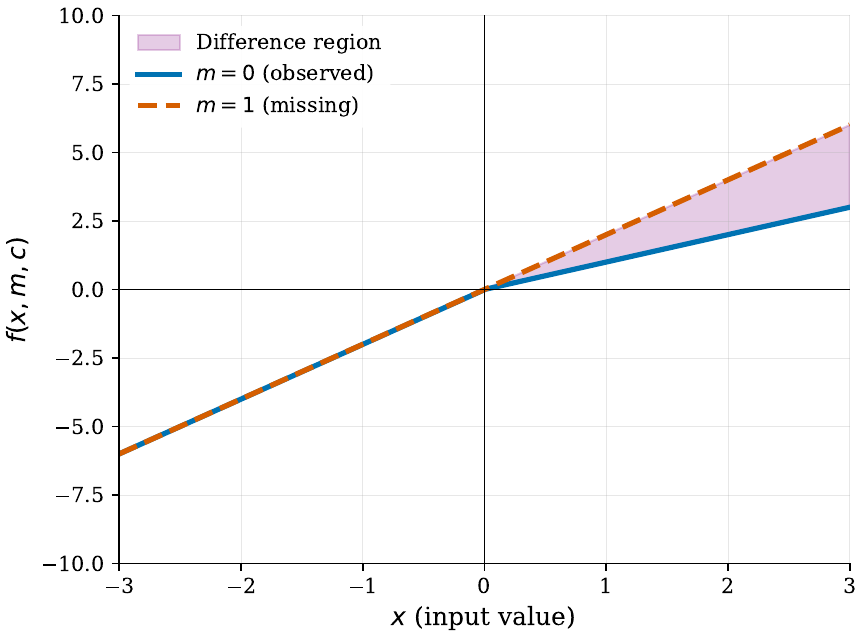}
    \vspace{-2ex}
    \caption{\small Evolved missing-aware activation functions (Glass dataset).}
    \Description{}
    \label{fig:Missing}
\end{figure}

To understand the behavior of the evolved activation functions, we analyze solutions found by the GP search and their behavior. Figure~\ref{fig:missingpima} presents an evolved activation function demonstrating three-channel awareness. The figure shows the confidence effect: the activation shape transforms from complex nonlinear response at high confidence ($c=1.0$) to conservative near-linear behavior at low confidence ($c=0.3$), while holding $m=0$. The evolved function $f(x,m,c) = x(c+2) \left[ \tfrac{c}{\mathrm{ReLU}(\max(c,-0.5)-m)} - c \, \mathrm{ReLU}(1.5cx)(0.5+c) \right]$
integrates both missing indicators and confidence scores, enabling adaptive behavior based on data reliability. Figure \ref{fig:Missing} shows a strong activation function that shows its awareness of the missing data. The function $f(x,m,c) = \min(x \cdot m, x) + x$ shows its conditional behavior based on the missing flag. For positive inputs, when the data is observed ($m=0$), the output is $x$, but when the data is imputed ($m=1$), the output is $2x$. This suggests that the evolutionary process has created a strong adaptation mechanism, where the function has learned to double the imputed data, possibly to counter the uncertainty of the imputed data. Table~\ref{tab:winners} shows the winner activation formulae that were developed over 30 independent GP runs on the Hepatitis dataset. The variety of solutions found shows how flexible the three-channel search space is. Although the majority of solutions include missingness or confidence channels, a number of runs (e.g., 4, 9, 16, 28) generate competitive activations using only $x$, suggesting that the GP preserves channels selectively based on validation performance rather than being forced into uniform three-channel solutions by the diversity incentive.

% The fact that missingness and confidence channels are used consistently between runs shows that the suggested method works. 

\begin{table}[htbp]
\centering
\caption{Winner Activation Functions Evolved Across 30 Independent GP Runs on the Hepatitis Dataset.}
\label{tab:winners}
\scriptsize
\renewcommand{\arraystretch}{0.82}
\setlength{\tabcolsep}{3pt}
\setlength{\abovecaptionskip}{2pt}
\setlength{\belowcaptionskip}{0pt}
    \begin{tabular}{c c c >{\raggedright\arraybackslash}p{0.62\columnwidth}}
    \toprule
    \textbf{Run} & \textbf{Acc.} & \textbf{F1} & \textbf{Evolved Formula $f(x,m,c)$} \\
    \midrule
    1  & 0.7742 & 0.8727 & $x^{2}\cdot c$ \\
    
    2  & 0.8387 & 0.9057 & $\min\!\left(\operatorname{softplus}(x) + \frac{x - 2c}{\operatorname{softplus}(m/0.1)},\, x\right)$ \\
    
    3  & 0.8387 & 0.9020 & $\tanh\!\left(\max\!\left(\min\!\left(\exp(\max(m,x)),-0.1\right),|x|\right)\right)+x$ \\
    
    4  & 0.7419 & 0.8462 & $\max\!\left(-1,\;\max(x,\sin x)\cdot\min\!\left(x,\frac{x}{\sin x}\right)\right)$ \\
    
    5  & 0.8065 & 0.8571 & $x\cdot\max(c,\operatorname{softplus}(-0.1))$ \\
    
    6  & 0.8065 & 0.8750 & $\frac{x}{0.5}-\max(m,c)$ \\
    
    7  & 0.7742 & 0.8444 & $(c-m-x)+\cos\!\left(x+m+c+0.5+\min(c,2-c)\right)-x$ \\
    
    8  & 0.9032 & 0.9388 & $\operatorname{ELU}(\exp(m+x))$ \\
    
    9  & 0.9032 & 0.9362 & $0.5\,\sin(x)$ \\
    
    10 & 0.8065 & 0.8750 & $x+\min\!\left(x,\,m-x^{2}\right)$ \\
    
    11 & 0.8387 & 0.8936 & $\sin(\min(x,0))+\max\!\left(c,\exp(x)-\exp\!\left(c+\operatorname{ELU}(\tanh(\exp x))\right)+x\right)+2x$\\
    
    12 & 0.6774 & 0.8000 & $\operatorname{ELU}\!\left(\operatorname{ReLU}\!\left(1-\frac{x}{c}+m-\operatorname{ReLU}(x)-m+x^{3}\right)\right)^{2}$ \\
    
    13 & 0.8065 & 0.8750 & $x\cdot\left(\min\!\left(x,\,x+c+\operatorname{LeakyReLU}(x)\right)-1\right)$ \\
    
    14 & 0.8065 & 0.8750 & $-\min\!\left(x,\min(c^{2},m)\right)$ \\
    
    15 & 0.8065 & 0.8750 & $(c-\max(\cos(2c),x))\cdot x$ \\
    
    16 & 0.7419 & 0.8400 & $\sin(x)$ \\
    
    17 & 0.7742 & 0.8727 & $\min(x,c)$ \\
    
    18 & 0.8065 & 0.8800 & $c\cdot\left(m-\min\!\left(-\max\!\left(x+m\cdot x+\operatorname{ReLU}(x+c)\right),0.5\right)\right)$ \\
    
    19 & 0.6774 & 0.7500 & $\frac{x}{c}$ \\
    
    20 & 0.7742 & 0.8627 & $\max(x^{2},m)$ \\
    
    21 & 0.7097 & 0.8302 & $\max(\exp m,\max(c,c^{2}))+\frac{0.5+\cos x}{x}$ \\
    
    22 & 0.7742 & 0.8627 & $(x)\cdot(c-m)$ \\
    
    23 & 0.7419 & 0.8462 & $x\cdot(x-(x-c)m)-x$ \\
    
    24 & 0.8387 & 0.8837 & $x+m$ \\
    
    25 & 0.8710 & 0.9167 & $\frac{\sqrt{|c|}+m-\sqrt c+0.1}{|c|}-\sqrt c-2x$ \\
    
    26 & 0.8387 & 0.8980 & $\min(m,x)-m\cdot\left(m-(x+\max(x-\!x,\sqrt{x}\sigma(c)))-x\right)$ \\
    
    27 & 0.8065 & 0.8750 & $\frac{\max(x^{2}+x,x)}{\tanh\!\left(0.1/\min(\max(-\operatorname{ELU}(\exp(\cos(\cos 2))x),x),m)\right)}$ \\
    
    28 & 0.7419 & 0.8462 & $x+x^{2}$ \\
    
    29 & 0.7097 & 0.8302 & $\max(\min(m,c),x)$ \\
    
    30 & 0.8387 & 0.9057 & $x-\max(x-m,-\min(x,2))$ \\
    
    \bottomrule
    \end{tabular}
    \vspace{-4pt}
\end{table}

\section{Conclusions and Future Work}

This paper proposed Three-Channel Evolved Activations (3C-EA), a GP method for evolving multivariate activation functions $f(x,m,c)$ that directly includes feature values, missingness indicators, and confidence scores. To effectively transmit missingness and confidence information through deep models, we proposed ChannelProp, a deterministic method for propagating missingness and confidence information through linear layers. Experiments on natural and introduced missing data (MCAR, MAR, MNAR) showed that 3C-EA is competitive with, and often better than standard fixed activation functions (ReLU, Swish, LeakyReLU, ELU). Experiments showed that improvements were consistent across different types of missingness, with significant improvements seen in datasets with higher missing rates and multi-class classification tasks. The study of evolved functions proved that GP can detect effective adaptive strategies, such as enhancing imputed values or modifying activation functions based on confidence levels.

Future work includes extending ChannelProp beyond fully connected layers to convolutional and attention-based architectures, where reliability propagation could operate across spatial or sequence dimensions, potentially enabling confidence-aware processing in CNNs, Transformers, and large language models. Another promising direction is to treat different  missingness mechanisms (MCAR, MAR, and MNAR) distinctly, for example, through additional flags and architectural components designed to respond differently to each mechanism. In addition, combining heuristic confidence 
scores with learned confidence estimates remains an open direction for further improvement.

%\FloatBarrier
\balance
\bibliographystyle{ACM-Reference-Format}
\bibliography{reference}

\end{document}